\documentclass[10pt,twocolumn,letterpaper]{article}

\usepackage[pagenumbers]{cvpr} 

\definecolor{cvprblue}{rgb}{0.21,0.49,0.74}
\usepackage[pagebackref,breaklinks,colorlinks,allcolors=cvprblue]{hyperref}

\usepackage[ruled,linesnumbered]{algorithm2e}
\usepackage{multirow}
\usepackage{booktabs}
\usepackage{diagbox}
\usepackage{colortbl}
\usepackage{tabularx}
\usepackage{ragged2e}
\usepackage{amsmath}
\usepackage{lipsum}
\usepackage{float}
\usepackage{amssymb}
\usepackage{color,xcolor}
\usepackage{float}

\def\paperID{3270} 
\def\confName{CVPR}
\def\confYear{2026}

\title{ScoreAdv: Score-based Targeted Generation of Natural Adversarial Examples via Diffusion Models}

\author{Chihan haung\textsuperscript{\rm 1,}
\\
Nanjing University of Science and Technology\\
{\tt\small huangchihan@njust.edu.cn}
\and
Hao Tang\thanks{Corresponding author.}\\
Peking University\\
{\tt\small haotang@pku.edu.cn}
}

\begin{document}
\maketitle
\begin{abstract}
Despite the success of deep learning across various domains, it remains vulnerable to adversarial attacks. Although many existing adversarial attack methods achieve high success rates, they typically rely on $\ell_{p}$-norm perturbation constraints, which do not align with human perceptual capabilities. Consequently, researchers have shifted their focus toward generating natural, unrestricted adversarial examples (UAEs). GAN-based approaches suffer from inherent limitations, such as poor image quality due to instability and mode collapse. Meanwhile, diffusion models have been employed for UAE generation, but they still rely on iterative PGD perturbation injection, without fully leveraging their central denoising capabilities. In this paper, we introduce a novel approach for generating UAEs based on diffusion models, named ScoreAdv. This method incorporates an interpretable adversarial guidance mechanism to gradually shift the sampling distribution towards the adversarial distribution, while using an interpretable saliency map to inject the visual information of a reference image into the generated samples. Notably, our method is capable of generating an unlimited number of natural adversarial examples and can attack not only classification models but also retrieval models. We conduct extensive experiments on ImageNet and CelebA datasets, validating the performance of ScoreAdv across ten target models in both black-box and white-box settings. Our results demonstrate that ScoreAdv achieves state-of-the-art attack success rates and image quality, while maintaining inference efficiency. Furthermore, the dynamic balance between denoising and adversarial perturbation enables ScoreAdv to remain robust even under defensive measures.
\end{abstract}

\section{Introduction}

Recent years have witnessed the remarkable application of deep learning in various domains, including image classification \cite{hao2023-reconciliation}, autonomous driving \cite{feng2023-dense}, medical image analysis \cite{zhang2023-tformer}, face recognition \cite{omkar2015-deep}, etc. However, studies such as those of \cite{christian2014-intriguing} have demonstrated that deep learning models are highly susceptible to adversarial examples, which can mislead them into making incorrect decisions, thus compromising their security in real-world applications. Adversarial examples refer to subtle, imperceptible perturbations added to clean examples that cause the target model to make erroneous predictions with high confidence while remaining undetectable to the human eye. The transferability of adversarial examples poses an even greater security risk, as adversarial samples crafted for one target model can be effectively transferred to different models, potentially compromising the security of multiple systems.

\begin{figure*}[h]
  \centering
  \includegraphics[width=0.9\linewidth]{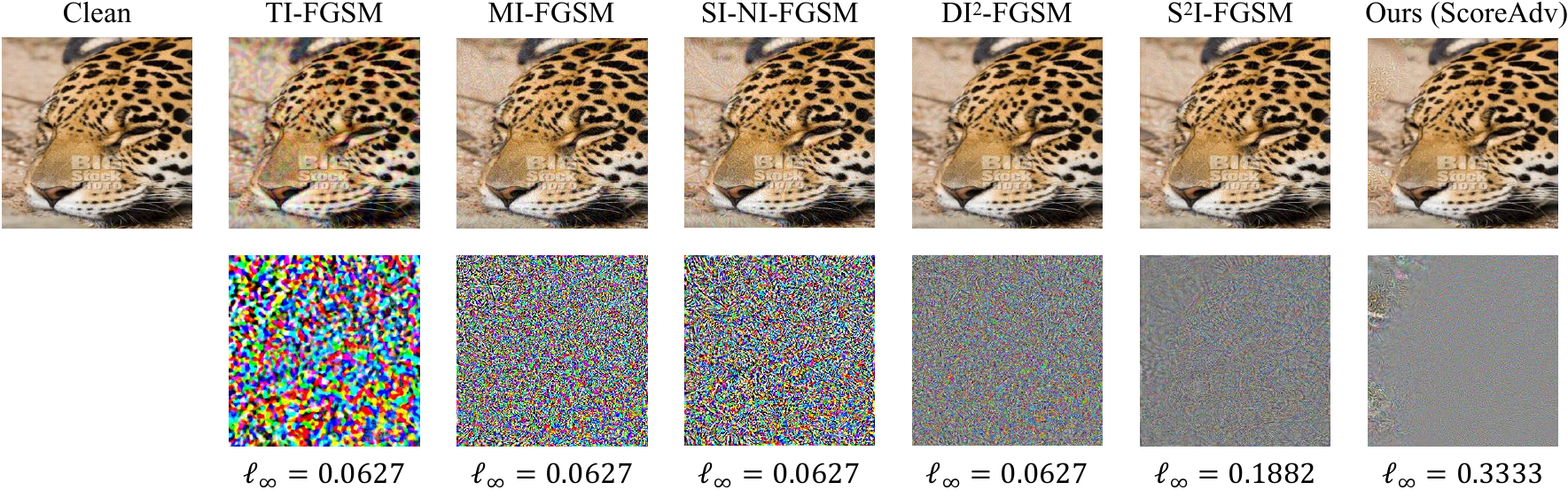}
  \caption{Adversarial attack results generated by various attack methods. The second row illustrates the difference between the benign and adversarial images. For detailed examination, please enlarge the image.}
  \label{l-restricted}
\end{figure*}

Many previous adversarial attacks have been based on $\ell_{p}$-norm constraints \cite{dong2019-evading, zhao2024-improving} to bound perturbation strength. However, recent research \cite{zhao2020-towards} has shown that $\ell_{p}$-norm poorly aligns with human perceptual similarity. As shown in Figure \ref{l-restricted}, attacks based on $\ell_{p}$-norm, despite having a small $\ell_{\infty}$ value, are more easily noticeable to the human eye.

In contrast to attacks based on $\ell_{p}$-norm perturbations, this paper investigates unrestricted adversarial example generation, where the generated samples are not constrained by any norm limitations, but instead optimize the input noise vector directly. Such perturbations typically exhibit large-scale, semantically coherent patterns that enhance attack success rates while remaining imperceptible \cite{jia2022-adv}.

The concept of UAE was first introduced by \cite{song2018-constructing}, who employed a generative adversarial network (GAN) with an auxiliary classifier to directly generate adversarial samples. However, GAN-based methods suffer from instability, mode collapse, and limited interpretability, often resulting in lower-quality outputs. In contrast, diffusion models \cite{ho2020-denoising} offer more stable training, better image quality, and broader distribution coverage \cite{dhariwal2021-diffusion}, making them more suitable for generating natural adversarial images.

The iterative denoising process in diffusion models naturally serves as a robust mechanism, capable of mitigating adversarial noise while retaining attack efficacy. \cite{dhariwal2021-diffusion} also highlighted that conditions can be incorporated into the denoising process to guide the generation. This characteristic enables us to carefully introduce adversarial information during the denoising process, thus achieving the generation of natural adversarial samples. However, existing diffusion-based attack methods have not fully exploited this capability. They often rely on PGD-based perturbations at each step \cite{chen2023-advdiffuser, chen2023-content}, or lack reference-image conditioning \cite{dai2025-advdiff}, limiting image quality and the full use of diffusion's potential.

Generating natural-looking adversarial examples from diffusion priors poses two linked challenges: (1) steering the stochastic sampling trajectory toward adversarial regions of the data manifold without destroying realism, and (2) doing so in a way that generalizes across classifiers and recognition systems under realistic constraints. 

In this paper, we propose ScoreAdv, a novel and interpretable UAE generation method based on diffusion models, which addresses both challenges by moving the adversarial mechanism inside the diffusion posterior instead of treating perturbation as an external add-on. Compared to previous diffusion-based adversarial attack methods, ScoreAdv \textbf{(I)} internalize stepwise, score-based adversarial guidance that reweights the denoising posterior at every sampling step, \textbf{(II)} fuse semantic structure via ScoreCAM-derived masks computed on a surrogate model to preserve fine-grained content, and \textbf{(III)} perform lightweight noise optimization to align high-variance latents with adversarial objectives without retraining the model. This unique combination yields a training-free, model-agnostic pipeline that attacks both classifiers and recognition systems, delivers strong black-box transfer, and keeps inference costs similar to the original diffusion model.

Our contributions can be summarized as follows:
\begin{enumerate}
\item We propose ScoreAdv, a training-free, unrestricted adversarial example generation framework based on diffusion models. It can generate an unlimited number of natural adversarial images, and its internalized score-based guidance yields a unified mechanism for both classification and recognition attacks while preserving realism.
\item We introduce an interpretable perturbation mechanism at each diffusion step to guide sampling, and incorporate interpretable saliency-based visual information from an optional reference image to improve semantic relevance.
\item Comprehensive experiments on ten target models demonstrate that ScoreAdv outperforms existing methods in both black-box and white-box settings, achieving superior attack success and visual quality, while keeping inference highly efficient. The dynamic interplay between denoising and adversarial perturbation also improves robustness against defenses.
\end{enumerate}

\section{Related Work}

\begin{figure*}[h]
  \centering
  \includegraphics[width=0.9\linewidth]{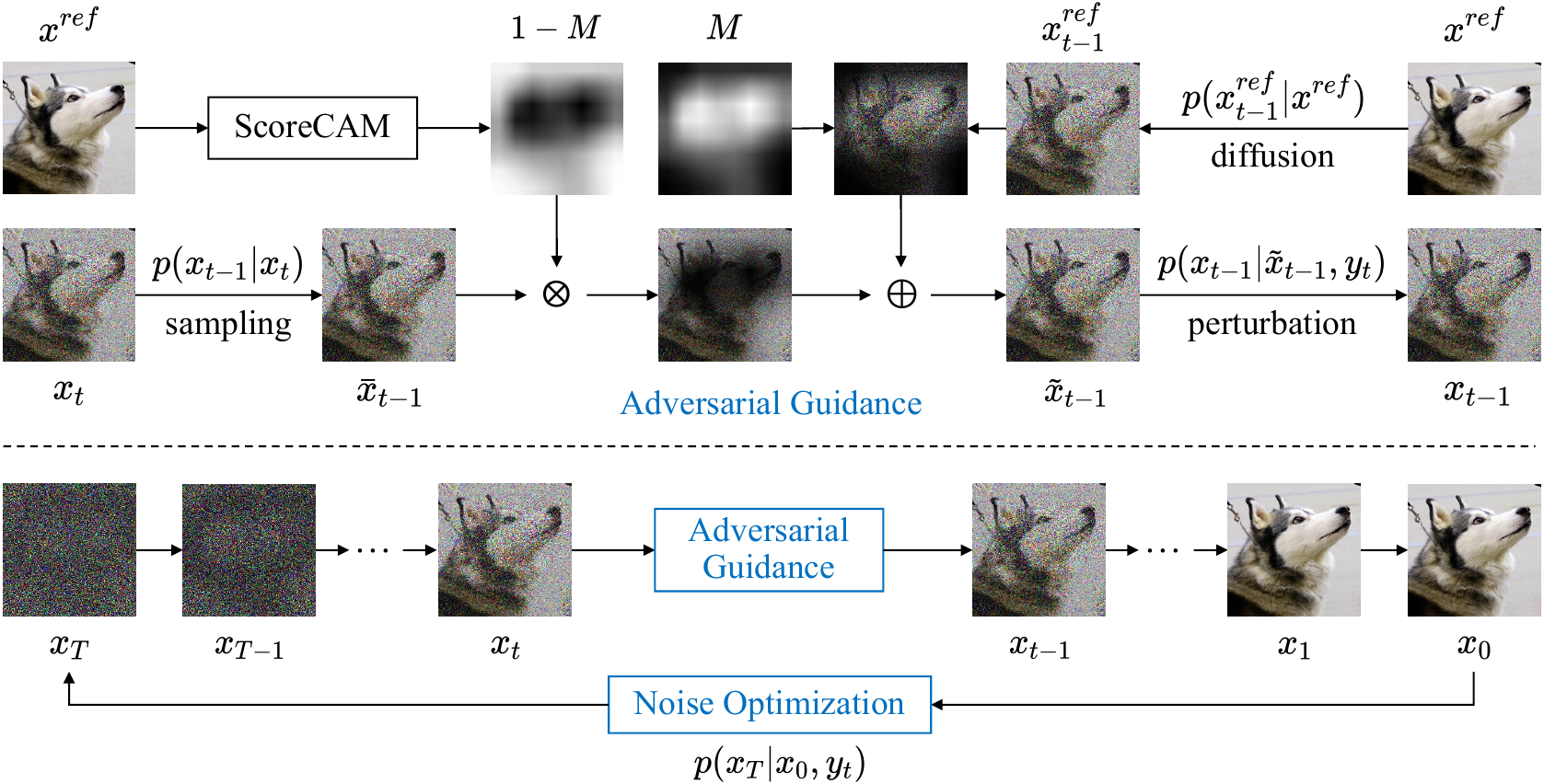}
  \caption{Overall framework of our ScoreAdv to generate adversarial images. The lower part illustrates the trained diffusion model. In each diffusion step, we employ adversarial guidance by first sampling $\bar{\boldsymbol{x}}_{t-1}$ using the diffusion model. Subsequently, adversarial perturbation is introduced based on the gradient derived from the target label and target model. If the content information from a reference image is required, we utilize ScoreCAM to generate its saliency map. We use the diffusion process to obtain $\boldsymbol{x}_{t-1}^{ref}$, which is then weighted and combined with $\tilde{\boldsymbol{x}}_{t-1}$ to produce the input for the next diffusion step $\boldsymbol{x}_{t-1}$.}
  \label{overall framework}
\end{figure*}

\subsection{Adversarial Attacks}

Adversarial attacks are commonly classified as white-box or black-box, depending on the attacker’s knowledge of the target model. White-box attacks assume full access to model architecture, gradients, and parameters, with representative methods including FGSM \cite{goodfellow2015-explaining}, PGD \cite{madry2018-towards}, and DeepFool \cite{moosavi2016-deepfool}. In contrast, black-box attacks, which only rely on input-output queries, are more challenging and better reflect real-world conditions. Notable examples include ZOO \cite{chen2017-zoo} and DeepSearch \cite{zhang2020-deepsearch}. Attacks can also be untargeted, aiming to misclassify into any incorrect label, or targeted, forcing misclassification into a specific class, often posing greater threats. Many black-box attacks leverage transferability \cite{micolas2016-transferability, dong2018-boosting}, applying white-box techniques on surrogate models to generate transferable adversarial examples.

\subsection{Unrestricted Adversarial Attacks}

Due to the inadequacy of $\ell_p$-norm in capturing perceptual similarity in RGB space, researchers have increasingly explored unrestricted adversarial attacks. Early works, such as \cite{zhao2018-generating}, modified GAN latent codes via auxiliary classifiers, but suffered from limited image quality. \cite{bhattad2019-unrestricted} perturbed color and texture, while \cite{yuan2022-natural} used a color distribution library and a neighborhood search to improve realism. \cite{jia2022-adv} generated attribute-based adversarial noise targeting facial features. More recent methods leverage diffusion models for unrestricted attacks. \cite{dai2025-advdiff} introduced adversarial guidance for stable generation; \cite{chen2023-advdiffuser} proposed AdvDiffuser, injecting PGD perturbations during sampling; and \cite{chen2025-diffusion} employed latent-space mapping with content-preserving structures and attention-based distractions to craft adversarial examples.

\subsection{Diffusion Models}

Diffusion models are based on stochastic processes and Markov chains, involving a forward process of gradually adding noise and a reverse denoising process. Given an image $\boldsymbol{x}_0$, a total of $T$ diffusion steps, and a noise schedule $\beta_{1:T}$, the forward process uses a fixed Markov chain to iteratively add Gaussian noise, producing a sequence of increasingly degraded images.
\begin{equation}
q(\boldsymbol{x}_t|\boldsymbol{x}_{t-1})=\mathcal{N}\left(\sqrt{1-\beta_t}\boldsymbol{x}_{t-1},\beta_t\mathbf{I}\right).
\end{equation}

In the denoising process, the diffusion model learns to predict and remove noise, effectively restoring $\boldsymbol{x}_T$ back to its original form $\boldsymbol{x}_0$.
\begin{equation}
p_\theta(\boldsymbol{x}_{t-1}|\boldsymbol{x}_t)=\mathcal{N}(\mu_\theta(\boldsymbol{x}_t,t),\sigma_\theta(\boldsymbol{x}_t,t)),
\end{equation}
where $\mu_{\theta}\left(\boldsymbol{x}_{t},t\right)=\frac{1}{\sqrt{\alpha_{t}}}\left(\boldsymbol{x}_{t}-\frac{\beta_{t}}{\sqrt{1-\bar{\alpha}_{t}}}\epsilon_{\theta}\left(\boldsymbol{x}_{t},t\right)\right)$, for simplification, $\alpha = 1 - \beta$, $\bar{\alpha}_t=\prod_{i=1}^t\alpha_i$, $\bar{\beta}_{t}=\frac{1-\bar{\alpha}_{t-1}}{1-\bar{\alpha}_{t}}\beta_{t}$.

Subsequent advances include Improved DDPM \cite{alex2021-improved}, which learns a variance schedule to improve sampling quality and efficiency. \cite{dhariwal2021-diffusion} introduced conditional diffusion, incorporating the gradient of cross-entropy loss from a target model into the sampling mean for class-conditioned generation.
\begin{equation}
p_{\theta,\phi}\left(\boldsymbol{x}_{t-1}| \boldsymbol{x}_t,y\right)=\mathcal{N}\left(\mu_\theta\left(\boldsymbol{x}_t\right)+\sigma_\theta\nabla_{\boldsymbol{x}_t}\log p_\phi\left(y| \boldsymbol{x}_t\right),\sigma_\theta\right).
\end{equation}

Furthermore, classifier-free guidance \cite{jonathan2021-classifier} enables the incorporation of class information without the need for an additional trained classifier.
\begin{equation}
\tilde{\boldsymbol{\epsilon}}_t=\left(1+s_c\right)\epsilon_\theta\left(\boldsymbol{x}_t,y\right)-s_c \epsilon_\theta(\boldsymbol{x}_t). 
\label{classifier-free}
\end{equation}

\section{Methodology}

\subsection{Formulation}

Instead of adding perturbations to clean samples, we aim to leverage a generative model to produce UAEs \cite{song2018-constructing}, yielding more natural adversarial inputs. Given a generative model $\mathcal{G}$, a target model $f$, ground truth label $y$, target label $y_{tar}$, and an optional reference image $\boldsymbol{x}^{ref}$, the generated UAE is defined as:
\begin{equation}
A_{UAE} \triangleq \{\boldsymbol{x}_{adv} \in \mathcal{G}(\boldsymbol{z}_{adv}, y, \boldsymbol{x}^{ref}) | f(\boldsymbol{x}_{adv}) = y_{tar} \}.
\end{equation}

The overall framework of the proposed ScoreAdv is shown in Figure \ref{overall framework}. In this paper, we focus on targeted attacks, where the goal is to generate a sample with true label $y$ that the target model misclassifies as a specified label $y_{tar}$.

\subsection{Adversarial Perturbation}

When using a pre-trained diffusion model, the denoising process gradually transforms noise into a natural image. Inspired by \cite{dhariwal2021-diffusion}, we found that during the denoising process, the gradient $\nabla_{\boldsymbol{x}_t}\log p_\phi(y|\boldsymbol{x}_t)$ can be used to guide the shift of the sampling distribution. Our objective is to generate an image $\boldsymbol{x}_0$ with a pre-trained diffusion model $\epsilon_\theta(\boldsymbol{x}_t, y)$, which appears consistent with the target label $y$, but is misclassified by the target model as $y_{tar} \neq y$. We achieve this through adversarial perturbations.

First, based on classifier-free guidance, we can use a pre-trained diffusion model to obtain the sampling formula guided by the ground truth label $y$ as the target class.
\begin{equation}
\boldsymbol{\bar{x}}_{t-1}=\frac{1}{\sqrt{\alpha_t}}\left(\boldsymbol{x}_t-\frac{1-\alpha_t}{\sqrt{1-\bar{\alpha}_t}}\tilde{\epsilon}_t\right) + \sigma_t z_t,
\end{equation}
where $\epsilon_{\theta}$ represents the pre-trained diffusion model, $s_c$ is the classifier guidance scale, $\tilde{\epsilon}_t$ is the predicted noise obtained using classifier-free guidance following \eqref{classifier-free}, $z_t$ is random noise, and $\bar{\boldsymbol{x}}_{t-1}$ is the image sampled with the true label $y$, incorporating class information.

Next, we aim to sample a sample near $\bar{\boldsymbol{x}}_{t-1}$ that embodies the semantic information of the target class $y_{tar}$:
\begin{equation}
p\left(\tilde{\boldsymbol{x}}_{t-1}|\bar{\boldsymbol{x}}_{t-1},f\left(\tilde{\boldsymbol{x}}_{t-1}\right) = y_{tar}\right).
\end{equation}

This sampling process is similar to that of conditional diffusion models. We can use the gradient information of $\bar{\boldsymbol{x}}_{t-1}$ from the target model to guide this sampling.
\begin{equation}
\tilde{\boldsymbol{x}}_{t-1}=\bar{\boldsymbol{x}}_{t-1} + \sigma_t^2 s_a\nabla_{\bar{\boldsymbol{x}}_{t-1}}\log p_f\left(y_{tar}|\bar{\boldsymbol{x}}_{t-1}\right),
\label{adversarial guidance}
\end{equation}
where $s_a$ represents the adversarial guidance scale.

The derivation of Equation \eqref{adversarial guidance} is provided in Supplementary Material \ref{eq8derivation}. Intuitively, a small perturbation is introduced to shift $\bar{\boldsymbol{x}}_{t-1}$ toward the target class $y_{tar}$, increasing the likelihood that the generated sample $\tilde{\boldsymbol{x}}_{t-1}$ is classified as $y_{tar}$.

\begin{algorithm}[t]
  \SetKwData{Left}{left}\SetKwData{This}{this}\SetKwData{Up}{up}
  \SetKwFunction{Union}{Union}\SetKwFunction{FindCompress}{FindCompress}
  \SetKwInOut{Input}{Input}\SetKwInOut{Output}{Output}
  \KwIn{Ground truth class label $y$, (optional) reference image $\boldsymbol{x}^{ref}$, target label $y_{tar}$ or target image $\boldsymbol{x}_{tar}$, target model $f$, pretrained diffusion model $\epsilon_{\theta}$, reverse generation process timestep $T$, noise optimization times $N$, adversarial guidance scale $s_a$, classifier guidance scale $s_c$, noise optimization guidance scale $s_n$, noise schedule $\beta_{1:T}$}
  \KwOut{Adversarial image $\boldsymbol{x}_{adv}$}
  \BlankLine
  $\boldsymbol{x}_{T} \sim \mathcal{N}\left(0, \boldsymbol{I}\right)$\;
  $\boldsymbol{x}_{adv} = \{\}$\;
  \For{$i=1, \ldots, N$}{
    \For{$t=T, \ldots, 1$}{
 	 \tcp{classifier-free sampling}
	 $\tilde{\boldsymbol{\epsilon}}_t=\left(1+s_c\right)\epsilon_\theta\left(\boldsymbol{x}_t,y\right)-s_c \epsilon_\theta(\boldsymbol{x}_t)$\;
	 $\boldsymbol{\bar{x}}_{t-1}=\frac{1}{\sqrt{\alpha_t}}\left(\boldsymbol{x}_t-\frac{1-\alpha_t}{\sqrt{1-\bar{\alpha}_t}}\tilde{\epsilon}_t\right) + \sigma_t z_t$\;
      \tcp{if use $\boldsymbol{x}^{ref}$}
	 \If{$\boldsymbol{x}^{ref}$}{
        $\boldsymbol{m} = \operatorname{ScoreCAM}\left(\boldsymbol{x}^{ref}, f, y\right)$\;
	   $\boldsymbol{x}_{t-1}^{ref}\sim\mathcal{N}\left(\sqrt{\overline{\alpha}_{t-1}}\boldsymbol{x}^{ref},\left(1-\overline{\alpha}_{t-1}\right)\boldsymbol{I}\right)$\;
	   $\boldsymbol{\bar{x}}_{t-1} = \boldsymbol{\tilde{x}}_{t-1}\odot\left(\boldsymbol{1}-\boldsymbol{m}\right) + \boldsymbol{x}_{t-1}^{ref} \odot \boldsymbol{m}$\;}
	 \tcp{adversarial gradient}
	 $\hat{\boldsymbol{x}}_0=\frac{1}{\sqrt{\bar{\alpha}_{t-1}}}\left(\boldsymbol{\bar{x}}_{t-1}-\sqrt{1-\bar{\alpha}_{t-1}}\cdot\tilde{\boldsymbol{\epsilon}}_{t-1}\right)$\;
	 \lIf{$y_{tar}$}{
	 $\boldsymbol{g} = \nabla_{\boldsymbol{x}_{t-1}}\log p_{f}\left(y_{tar}|\boldsymbol{\bar{x}}_{t-1}\right)$}
	 \lElse{$\boldsymbol{g} = \nabla_{\boldsymbol{\hat{\boldsymbol{x}}_0}}\log p_{f}\left(\boldsymbol{x}_{tar}|\hat{\boldsymbol{x}}_0\right)$}
 	 $\boldsymbol{\tilde{x}}_{t-1}=\boldsymbol{\bar{x}}_{t-1}+\sigma_{t}^{2}s_a \boldsymbol{g}$\;
    }
  \tcp{noise gradient}
  \lIf{$y_{tar}$}{
  $\boldsymbol{g}_n = \nabla_{\boldsymbol{x}_0}\log p_f\left(y_{tar}|\boldsymbol{x}_0\right)$}
  \lElse{$\boldsymbol{g}_n = \nabla_{\boldsymbol{x}_0}\log p_f\left(\boldsymbol{x}_{tar}|\boldsymbol{x}_0\right)$}
  $\boldsymbol{x}_{T} \leftarrow \boldsymbol{x}_{T} + \bar{\sigma}_{T}^{2} s_n \boldsymbol{g}_n$\;
  \tcp{reserve if attack successful}
  \lIf{$f(\boldsymbol{x}_0)=y_{tar}$}{$\boldsymbol{x}_{adv} \leftarrow \boldsymbol{x}_0$}
  }
  \caption{ScoreAdv Diffusion Sampling}\label{algorithm}
\end{algorithm}

\paragraph{Image Recognition.}
While the above adversarial perturbation targets classification tasks, our method also extends to image recognition. Given a target image $\boldsymbol{x}_{tar}$ and a similarity model $f(\cdot, \cdot)$, we apply the same procedure to obtain $\bar{\boldsymbol{x}}_{t-1}$, and then sample $\tilde{\boldsymbol{x}}_{t-1}$ as follows:
\begin{equation}
\begin{aligned}
\tilde{\boldsymbol{x}}_{t-1} &=\bar{\boldsymbol{x}}_{t-1} + \sigma_t^2 s_a\nabla_{\boldsymbol{\hat{\boldsymbol{x}}_0}}\log p_{f}\left(\boldsymbol{x}_{tar}|\hat{\boldsymbol{x}}_0\right), \\
\hat{\boldsymbol{x}}_0 &=\frac{1}{\sqrt{\bar{\alpha}_{t-1}}}\left(\bar{\boldsymbol{x}}_{t-1}-\sqrt{1-\bar{\alpha}_{t-1}}\cdot\tilde{\boldsymbol{\epsilon}}_{t-1}\right),
\end{aligned}
\label{recognition}
\end{equation}
where $\hat{\boldsymbol{x}}_0$ is the predicted clean image from $\bar{\boldsymbol{x}}_{t-1}$, and $\tilde{\boldsymbol{\epsilon}}_{t-1}$ is the predicted noise at step $t-1$. We measure similarity between $\boldsymbol{x}_{tar}$ and $\hat{\boldsymbol{x}}_0$ rather than $\bar{\boldsymbol{x}}_{t-1}$, since the latter contains significant noise. As the target model is trained on clean images, evaluating on denoised predictions better reflects its behavior. Thus, we first predict $\hat{\boldsymbol{x}}_0$ from $\bar{\boldsymbol{x}}_{t-1}$, then measure $f(\boldsymbol{x}_{tar} | \hat{\boldsymbol{x}}_0)$. Derivation is in Supplementary Material \ref{eq9derivation}

\subsection{ScoreCAM-based Inpainting}

We introduce ScoreCAM to guide the diffusion denoising process with a reference image, generating natural adversarial examples that resemble it. Unlike gradient-based methods such as GradCAM \cite{selvaraju2017-gradcam}, which suffer from vanishing gradients and instability \cite{wang2020-scorecam}, ScoreCAM avoids gradient dependence.

\begin{figure}[h]
  \centering
  \includegraphics[width=0.9\linewidth]{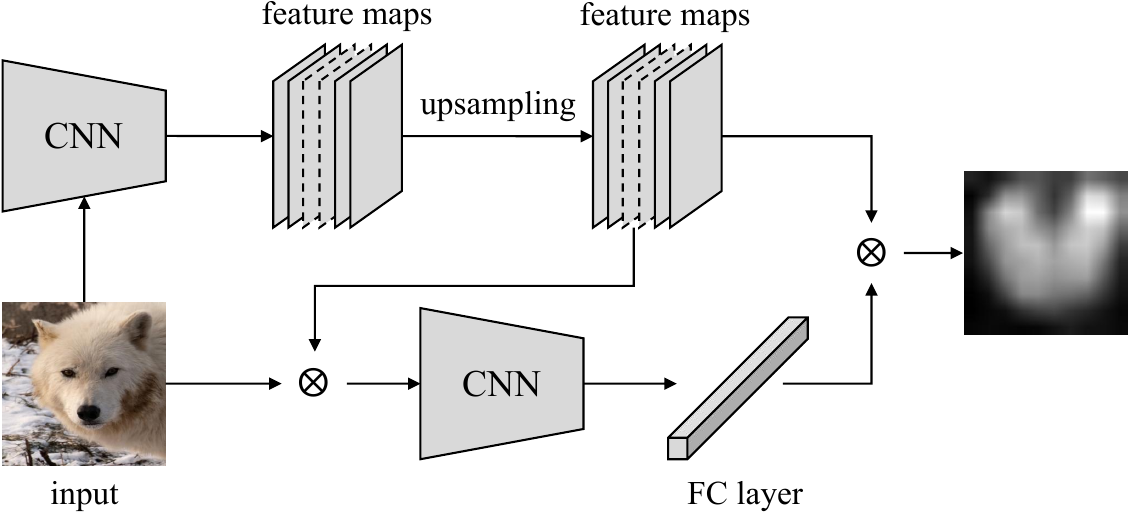}
  \caption{Illustration of ScoreCAM.}
  \label{scorecam}
\end{figure}

As shown in Figure \ref{scorecam}, given an image $X$ and model $f$ with layer $l$, we extract activation maps $A_l^k$ for each channel $k$ and assign weights based on the score increase $S$, yielding a more stable and interpretable saliency map.
\begin{equation}
S\left(A_l^k\right)=f\left(X \odot s\left(\operatorname{Up}\left(A_l^k\right)\right)\right),
\end{equation}
where $\operatorname{Up}\left(\cdot\right)$ represents the upsampling function, which upscales $A_l^k$ to the size of input $X$, and $s(x)=\frac{x-\min x}{\max x-\min x}$ is the normalization function that maps $x$ to [0, 1].

Thus, for a given class $c$, ScoreCAM can be computed as:
\begin{equation}
\begin{aligned}
\operatorname{ScoreCAM}^c & =\operatorname{ReLU} \left(\sum_k \alpha_k^c A_l^k \right), \\
\alpha_k^c & =\frac{\operatorname{exp}\left(S^c\left(A_l^k\right)\right)}{\displaystyle\sum_j \operatorname{exp}\left(S^c\left(A_l^j\right)\right)},
\end{aligned}
\end{equation}
where $S^c\left(\cdot \right)$ represents the logits corresponding to class $c$.

Thus, given the reference image $\boldsymbol{x}^{ref}$, target model $f$ and ground truth label $y$, we can obtain the saliency map $\boldsymbol{m}$.
\begin{equation}
\boldsymbol{m} = \operatorname{ScoreCAM}(\boldsymbol{x}^{ref}, f, y).
\end{equation}

At each diffusion step $t$, we gradually inject information from the reference image $\boldsymbol{x}^{ref}$ into $\boldsymbol{\tilde{x}}t$ by applying the forward diffusion process to obtain the noisy version $\boldsymbol{x}_{t-1}^{ref}$.
\begin{equation}
\boldsymbol{x}_{t-1}^{ref}\sim\mathcal{N}\left(\sqrt{\overline{\alpha}_{t-1}}\boldsymbol{x}^{ref},\left(1-\overline{\alpha}_{t-1}\right)\boldsymbol{I}\right).
\end{equation}

We leverage the concept of image inpainting \cite{lugmayr2022-repaint} to perform a weighted fusion of $\boldsymbol{x}_{t-1}^{ref}$ with $\boldsymbol{\tilde{x}}_t$.
\begin{equation}
\boldsymbol{x}_{t-1} = \boldsymbol{\tilde{x}}_{t-1}\odot\left(\boldsymbol{1}-\boldsymbol{m}\right) + \boldsymbol{x}_{t-1}^{ref} \odot \boldsymbol{m}.
\end{equation}

\subsection{Noise Optimization}

\cite{zhou2025-golden} show that certain noise types yield better results. Motivated by this, we incorporate prior information about the target label into the initial noise. Specifically, after generating $\boldsymbol{x}_0$, we apply diffusion process and use gradient signals from the target model to guide noise optimization.
\begin{equation}
\boldsymbol{x}_T= \left(\sqrt{\bar{\alpha}_t} \boldsymbol{x}_0 + \sqrt{1-\bar{\alpha}_t} \boldsymbol{\epsilon}\right) + \bar{\sigma}_T^2 s_n \nabla_{\boldsymbol{x}_0} \log p_f(y_{tar}|\boldsymbol{x}_0),
\label{noise optimization}
\end{equation}
where $\epsilon \sim \mathcal{N}\left(0, \boldsymbol{I}\right)$, the former represents the forward diffusion process, while the latter denotes the gradient of $\boldsymbol{x}_0$ with respect to the target model.

For image recognition models, the equation for noise optimization is similar to the one described above:
\begin{equation}
\boldsymbol{x}_T= \left(\sqrt{\bar{\alpha}_t} \boldsymbol{x}_0 + \sqrt{1-\bar{\alpha}_t} \boldsymbol{\epsilon}\right) + \bar{\sigma}_T^2 s_n \nabla_{\boldsymbol{x}_0} \log p_f(\boldsymbol{x}_{tar}|\boldsymbol{x}_0).
\end{equation}

Full derivation is in Supplementary Material \ref{eq15derivation}.

\subsection{ScoreAdv Algorithm}

Algorithm \ref{algorithm} summarizes the ScoreAdv pipeline. At each diffusion step, we sample $\bar{\boldsymbol{x}}_{t-1}$ using classifier-free guidance, then apply adversarial guidance to obtain $\tilde{\boldsymbol{x}}_{t-1}$. If a reference image $\boldsymbol{x}^{ref}$ is available, we use ScoreCAM to inject content information. After denoising for $\boldsymbol{x}_0$, noise optimization is performed, and the process is repeated for $N$ cycles.

\section{Experiments}

\subsection{Datasets}

\begin{table*}[!h]
\caption{Attack performance comparison of various methods on the ImageNet dataset. We report ASR (\%) and transferability. S. represents the surrogate model, while T. represents the target model. In cases where the surrogate model and the target model are identical, the scenario is classified as a white-box attack, with the background highlighted in gray for clarity. Avg. represents the average ASR of black-box attacks. The best result is \textbf{bolded} and the second best result is \underline{underlined}.}
\label{attack}
\small
\begin{tabularx}{\textwidth}{XlXXXllXllXX|X}
\midrule
\multirow{2}{*}{\diagbox[innerwidth=3em, height=2.5em]{S.}{T.}}   & \multirow{2}{*}{Attacks} & \multicolumn{5}{c}{CNNs}                            & \multicolumn{3}{c}{Transformers} & \multicolumn{2}{c}{MLPs} & \multirow{2}{*}{Avg.$\uparrow$} \\ \cmidrule(r){3-7} \cmidrule(r){8-10} \cmidrule(r){11-12}
                        &                          & Res50 & VGG19 & Inc-v3 & WRN50 & CNX & ViT-B    & Swin-B    & DeiT-B    & Mix-B       & Mix-L      &                       \\ \midrule
\multirow{9}{*}{Res-50} & PGD         & \cellcolor{lightgray!60}94.4   & 27.6  & 30.0   & 15.7           & 12.3     & 13.3     & 22.1      & 18.0      & 41.4        & 33.9       & 23.81 \\
                        & DI$^2$-FGSM    & \cellcolor{lightgray!60}91.0   & 23.3  & 22.5   & 25.6           & 12.7     & 9.9      & 14.4      & 11.2      & 42.1        & 40.0       & 22.41  \\
                        & S$^2$I-FGSM    & \cellcolor{lightgray!60}\underline{95.7}   & 29.5  & 31.9   & 28.6           & 17.9     & 16.1     & 23.8      & 15.8      & 45.8        & 44.3       & 28.19 \\
                        & cAdv        & \cellcolor{lightgray!60}92.5   & 45.0  & 30.0   & 25.7           & 19.3     & 20.3     & 27.8      & 21.0      & 51.4        & 49.9       & 32.27 \\
                        & NCF         & \cellcolor{lightgray!60}91.4   & 46.7  & 38.8   & 34.7           & 24.9     & 24.4     & 26.8      & 27.3      & 55.2        & 53.0       & 36.87 \\
                        & AdvDiffuser & \cellcolor{lightgray!60}91.4   & 50.9  & 43.3   & 39.2           & 40.8     & 37.5     & 46.7      & 42.0      & 56.4        & 53.8       & 45.62 \\
                        & DiffAttack  & \cellcolor{lightgray!60}87.0   & 51.2  & 42.1   & 40.7           & 41.0     & 35.9     & 45.0      & 41.1      & 58.0        & 55.3       & 45.59 \\
                        & Advdiff     & \cellcolor{lightgray!60}92.1   & \underline{53.0}  & \textbf{47.2}   & \underline{44.5}           & \underline{46.7}     & \textbf{40.4}     & \underline{46.8}      & \underline{45.8}      & \underline{62.4}        & \underline{60.1}       & \underline{49.66} \\ \cmidrule(r){2-13}
                        & ScoreAdv    & \cellcolor{lightgray!60}\textbf{97.6}   & \textbf{57.0}  & \underline{46.7}   & \textbf{48.9}           & \textbf{46.9}     & \underline{40.1}     & \textbf{47.0}      & \textbf{46.2}      & \textbf{65.3}        & \textbf{62.7}       & \textbf{51.20} \\ \midrule
\multirow{9}{*}{VGG19}  & PGD          & 20.6 & \cellcolor{lightgray!60}97.5  & 24.9   & 18.2           & 19.0     & 11.5     & 16.9      & 13.3      & 33.3        & 29.1       & 20.76 \\
                        & DI$^2$-FGSM    & 23.0 & \cellcolor{lightgray!60}\underline{98.2}  & 29.5   & 24.6           & 24.1     & 13.4     & 22.5      & 14.2      & 39.8        & 34.3       & 25.04 \\
                        & S$^2$I-FGSM     & 30.1 & \cellcolor{lightgray!60}\textbf{100.0} & 35.6   & 26.1           & 31.5     & 17.0     & 26.7      & 21.4      & 45.3        & 41.2       & 30.54 \\
                        & cAdv        & 36.5 & \cellcolor{lightgray!60}90.2  & 39.2   & 30.7           & 22.7     & 25.2     & 36.0      & 24.7      & 45.4        & 44.8       & 33.91 \\
                        & NCF         & 36.8 & \cellcolor{lightgray!60}90.8  & 41.1   & 30.3           & 25.1     & 26.6     & 37.7      & 29.5      & 49.4        & 46.1       & 35.84 \\
                        & AdvDiffuser & 40.7 & \cellcolor{lightgray!60}94.0  & 46.2   & 38.4           & 36.2     & 28.6     & 34.4      & 30.1      & 58.9        & 51.6       & 40.57 \\
                        & DiffAttack  & \underline{42.0} & \cellcolor{lightgray!60}93.4  & 45.3   & 38.0           & 30.0     & 28.5     & 35.7      & 30.5      & 55.6        & 48.2       & 39.31 \\
                        & Advdiff     & 41.8 & \cellcolor{lightgray!60}94.7  & \underline{48.5}   & \underline{40.1}           & \underline{37.4}     & \textbf{35.4}     & \underline{39.5}      & \underline{36.0}      & \underline{60.3}        & \underline{54.4}       & \underline{43.71} \\ \cmidrule(r){2-13}
                        & ScoreAdv    & \textbf{48.7} & \cellcolor{lightgray!60}\textbf{100.0} & \textbf{49.1}   & \textbf{47.2}           & \textbf{39.1}     & \underline{32.9}     & \textbf{41.6}      & \textbf{39.1}      & \textbf{85.7}        & \textbf{75.0}       & \textbf{50.93} \\ \midrule
\multirow{9}{*}{Inc-v3} & PGD         & 21.3 & 23.6 & \cellcolor{lightgray!60}\textbf{94.2}   & 15.3           & 14.7     & 13.2     & 15.9      & 15.2      & 26.5        & 24.1       & 18.87 \\
                        & DI$^2$-FGSM    & 20.0 & 28.5 & \cellcolor{lightgray!60}90.2   & 14.3           & 15.1     & 13.8     & 19.5      & 14.6      & 31.1        & 26.5       & 20.38 \\
                        & S$^2$I-FGSM    & 24.1 & 32.2 & \cellcolor{lightgray!60}93.1   & 25.7           & 27.1     & 19.8     & 25.3      & 21.2      & 39.2        & 35.4       & 27.78 \\
                        & cAdv        & 31.7 & 38.6 & \cellcolor{lightgray!60}83.3   & 27.2           & 26.2     & 29.6     & 34.2      & 30.7      & 43.5        & 42.6       & 33.81 \\
                        & NCF         & 35.5 & 39.3 & \cellcolor{lightgray!60}86.3   & 31.7           & 28.9     & 31.9     & 33.9      & 33.3      & 44.7        & 41.9       & 35.68 \\
                        & AdvDiffuser & 39.4 & 45.8 & \cellcolor{lightgray!60}89.1   & 36.6           & 34.6     & 34.4     & 40.2      & 37.6      & 53.1        & 50.8       & 41.39 \\
                        & DiffAttack  & 40.1 & 46.4 & \cellcolor{lightgray!60}88.5   & 35.5           & 34.8     & 35.3     & 41.3      & 38.1      & 50.7        & 48.7       & 41.21 \\
                        & Advdiff     & \textbf{42.9} & \underline{47.6} & \cellcolor{lightgray!60}91.4   & \underline{39.3}           & \underline{36.7}     & \textbf{39.9}     & \underline{44.2}      & \underline{41.8}      & \underline{54.5}        & \underline{52.3}       & \underline{44.36} \\ \cmidrule(r){2-13} 
                        & ScoreAdv    & \underline{40.6} & \textbf{67.2} & \cellcolor{lightgray!60}\underline{93.4}   & \textbf{39.9}           & \textbf{40.8}     & \underline{39.4}     & \textbf{45.1}      & \textbf{43.6}      & \textbf{72.9}        & \textbf{58.5}       & \textbf{49.78} \\ \midrule
\multirow{9}{*}{ViT-B}  & PGD         & 25.6 & 28.0 & 26.6 & 19.7 & 18.7 & \cellcolor{lightgray!60}86.4     & 28.7      & 22.5      & 29.0        & 26.2       & 25.00  \\
                        & DI$^2$-FGSM    & 24.9 & 35.0 & 30.6 & 26.5 & 22.8 & \cellcolor{lightgray!60}89.8     & 31.0      & 28.2      & 33.1        & 30.9       & 29.22 \\
                        & S$^2$I-FGSM    & 29.2 & 38.5 & 35.9 & 27.6 & 29.4 & \cellcolor{lightgray!60}\textbf{90.3}     & 34.5      & 31.0      & 41.7        & 38.2       & 34.00 \\
                        & cAdv        & 33.6 & 42.0 & 39.0 & 30.3 & 28.2 & \cellcolor{lightgray!60}85.8     & 35.9      & 32.4      & 45.6        & 41.2       & 36.47 \\
                        & NCF         & 40.6 & 44.9 & 43.1 & 33.8 & 33.3 & \cellcolor{lightgray!60}81.0     & 38.7      & 35.7      & 51.1        & 46.5       & 40.86 \\
                        & AdvDiffuser & 46.2 & 47.8 & 49.0 & 44.1 & 39.7 & \cellcolor{lightgray!60}87.6     & 42.4      & 40.1      & 54.6        & 51.4       & 46.14 \\
                        & DiffAttack  & 45.4 & 49.7 & 48.6 & \underline{44.3} & 38.5 & \cellcolor{lightgray!60}86.0     & 41.2      & 38.8      & 55.1        & 52.6       & 46.02 \\
                        & Advdiff     & \underline{50.6} & \underline{53.2} & \underline{52.0} & \textbf{46.4} & \textbf{43.7} & \cellcolor{lightgray!60}\underline{90.1}     & \underline{47.1}      & \underline{44.3}      & \underline{58.7}        & \underline{56.9}       & \underline{50.32} \\ \cmidrule(r){2-13}
                        & ScoreAdv    & \textbf{54.2}       & \textbf{57.3}      & \textbf{53.2}       & 42.8               & \underline{42.0}         & \cellcolor{lightgray!60} 88.7        & \textbf{57.9}          & \textbf{59.0}          & \textbf{85.2}            & \textbf{64.5}           & \textbf{57.34}                      \\ \midrule  
\end{tabularx}
\end{table*}

For image classification, we use 1,000 samples from the ImageNet validation set, covering 1,000 categories. Images are resized to $224 \times 224$ to match the diffusion model.

For face recognition, we conduct experiments on CelebA and LFW. CelebA contains 202,599 images across 10,177 identities, and LFW includes 13,233 images of 5,749 individuals. We randomly select 1,000 images for evaluation.

\subsection{Experiment Settings}

We utilize the pre-trained Guided Diffusion model from OpenAI \cite{dhariwal2021-diffusion}, avoiding training from scratch. We select a diffusion step as $T=1000$, noise optimization iterations as $N=3$, adversarial guidance scale as $s_a=0.3$, and noise optimization scale as $s_n=0.8$.

We evaluate attacks on models from three architecture families. CNN-based models include ResNet (Res50) \cite{he2016-deep}, VGG19 \cite{karen2015-very}, Inception-v3 \cite{christian2016-rethinking}, WideResNet50-2 (WRN50) \cite{zagoruyko2016-wide}, and ConvNeXt (CNX) \cite{liu2022-a}. Transformers models include ViT-B/16 (ViT-B) \cite{alexander2021-an}, Swin Transformer (Swin-B) \cite{liu2021-swin}, and DeiT-B \cite{hugo2020-training}. MLP-based models include Mixer-B/16 (Mix-B) and Mixer-L/16 (Mixer-L) \cite{ily2021-mlp}.

We use attack success rate (ASR) to measure attack effectiveness. To assess imperceptibility and visual quality, we report FID \cite{martin2017-gans}, LPIPS \cite{zhang2018-the}, PSNR, and SSIM \cite{hore2010-image}, which are all computed between the generated adversarial images and the reference images.

\subsection{Results}

\begin{figure*}[h]
  \centering
  \includegraphics[width=1.0\linewidth]{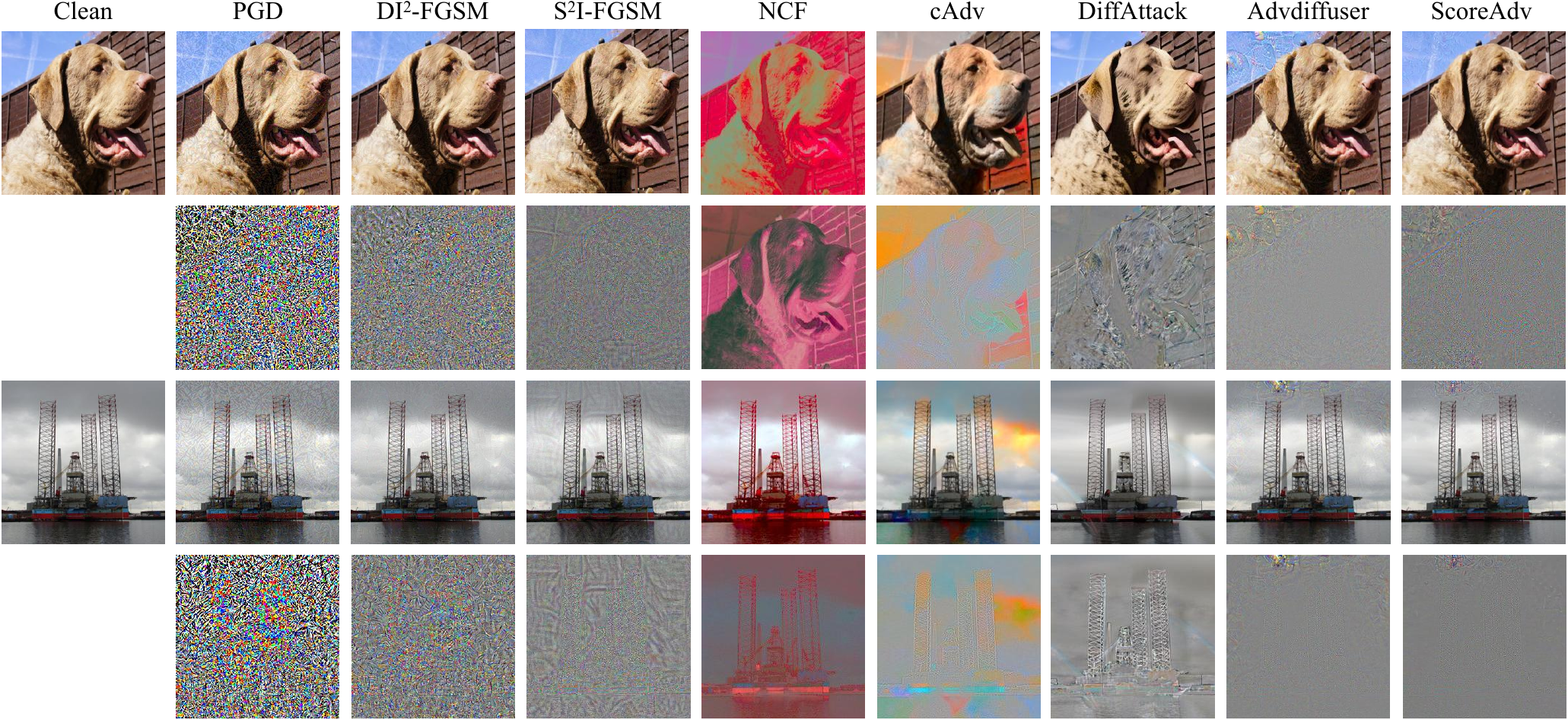}
  \caption{Visualization of adversarial images and perturbations generated by different attack methods. The upper row displays the adversarial images, while the lower row illustrates the corresponding perturbations.}
  \label{visual}
\end{figure*}

\begin{table*}[h]
\caption{Attack robustness comparison of various methods against different defense strategies. For SR, NRP, and DiffPure, we present the reduction in ASR following their application. The best result is \textbf{bolded} and the second best result is \underline{underlined}.}
\label{robustness}
\begin{tabularx}{\textwidth}{lXXXlll|llXX}
\midrule
  & AdvProp & R\&P & RandS   & Bit-Red & Adv-Inc-v3 & IncRes-v2$_{ens}$ & Inc-v3 & SR    & NRP   & Diffpure \\ \midrule
PGD         & 59.5 & 15.9 & 36.7 & 17.3 & 30.8       & 19.0         & \textbf{94.2}   & -26.1 & -49.6   & -51.2       \\
DI$^2$-FGSM    & 56.7 & 20.6 & 33.4 & 19.0 & 25.7       & 12.8         & 90.2   & -34.2 & -37.0   & -45.7       \\
S$^2$I-FGSM    & 75.2 & 22.7 & 54.9 & 21.8 & 41.0       & 28.0         & 93.1   & \textbf{-13.3} & \textbf{-25.2}   & -31.6       \\
cAdv        & 75.6 & 19.4 & 48.5 & 19.3 & 38.7       & 26.2         & 83.3   & -38.0 & -30.4   & -43.3       \\
NCF         & 78.2 & 19.1 & 50.6 & 17.5 & 42.5       & 27.8         & 86.3   & -21.5 & -27.8     & -33.8     \\
AdvDiffuser & 74.9 & 20.8 & 53.5 & 22.9 & 46.2       & 32.9         & 89.1   & -25.0 & -29.6    & -33.4      \\
DiffAttack  & 76.5 & 24.5 & 54.8 & 23.7 & 44.8       & 31.6         & 88.5   & -32.9 & -42.2    & -34.1      \\
Advdiff     & \underline{81.0} & \underline{49.1} & \underline{71.2} & \underline{44.4} & \underline{56.1}       & \underline{44.9}         & 91.4   & -18.4 & -28.6   & \underline{-20.4}      \\ \midrule
ScoreAdv    & \textbf{83.6} & \textbf{52.4} & \textbf{73.8} & \textbf{46.3} & \textbf{57.6} & \textbf{48.1} & \underline{93.4} & \underline{-17.6} & \underline{-25.3}    & \textbf{-18.7}      \\ \midrule
\end{tabularx}
\end{table*}

\paragraph{Targeted Attack Performance.}
Table \ref{attack} presents the ASR ofdifferent methods on the ImageNet dataset. We compare ScoreAdv with previous advanced adversarial attack methods such as PGD \cite{aleksander2017-towards}, DI$^2$-FGSM \cite{xie2019-improving}, S$^2$I-FGSM \cite{long2022-frequency}, cAdv \cite{bhattad2019-unrestricted}, NCF \cite{yuan2022-natural}, AdvDiffuser \cite{chen2023-advdiffuser}, DiffAttack \cite{chen2025-diffusion}, and Advdiff \cite{dai2025-advdiff}. Apart
from potential resolution adjustments, all implementations remain consistent with their settings.

As shown in Table \ref{attack}, ScoreAdv achieves SOTA performance in nearly all black-box and white-box scenarios. While ScoreAdv may slightly underperform in certain cases, these methods often yield lower-quality adversarial examples with poor black-box transferability. Additional results on recognition models are provided in Supplementary Material \ref{face performance}. Both quantitative and qualitative results show that ScoreAdv performs even better in this setting.

\paragraph{Generation Quality.}
\label{4.3.2}

\begin{figure*}[h]
	\includegraphics[width=0.247\linewidth]{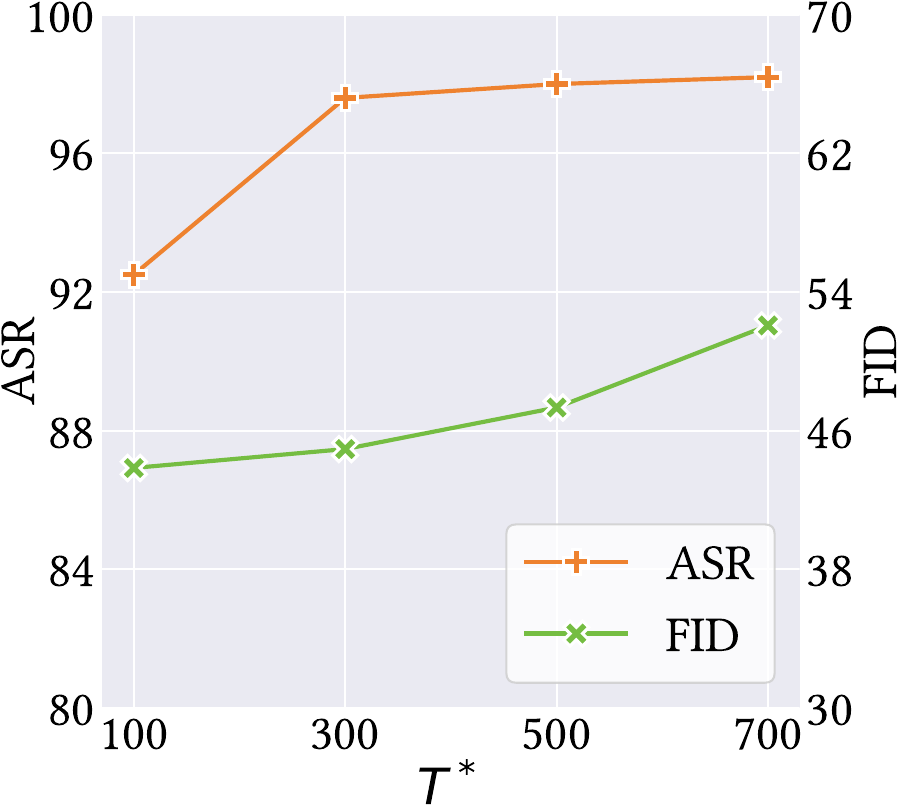}
	\includegraphics[width=0.247\linewidth]{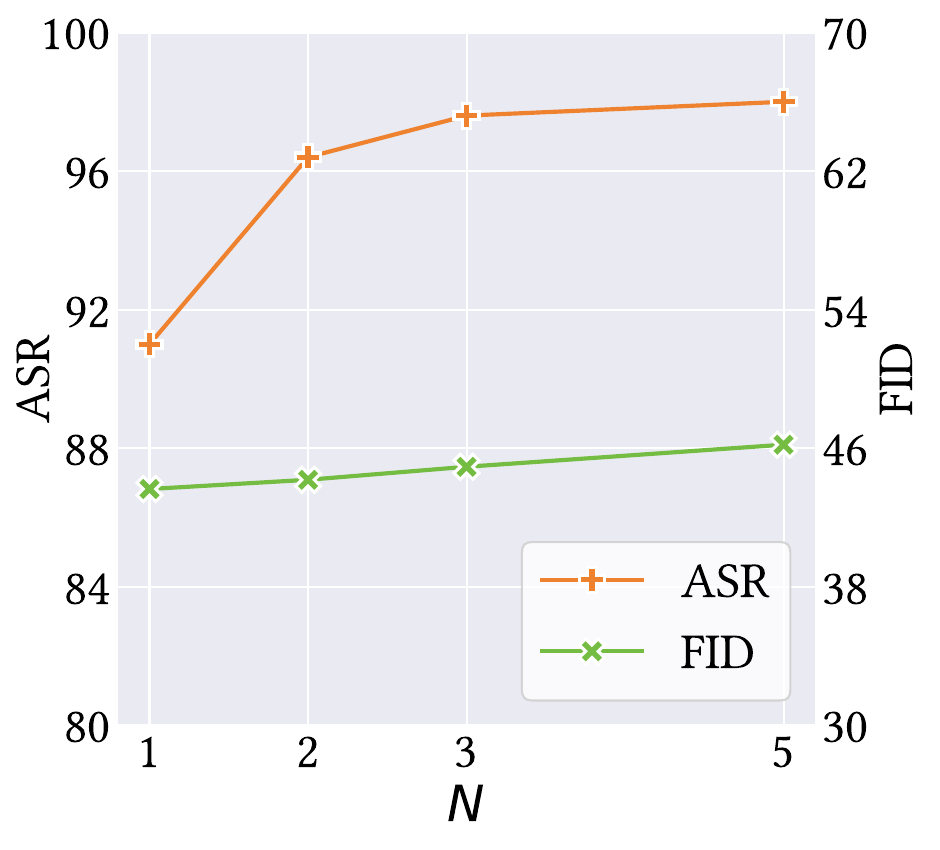}
	\includegraphics[width=0.247\linewidth]{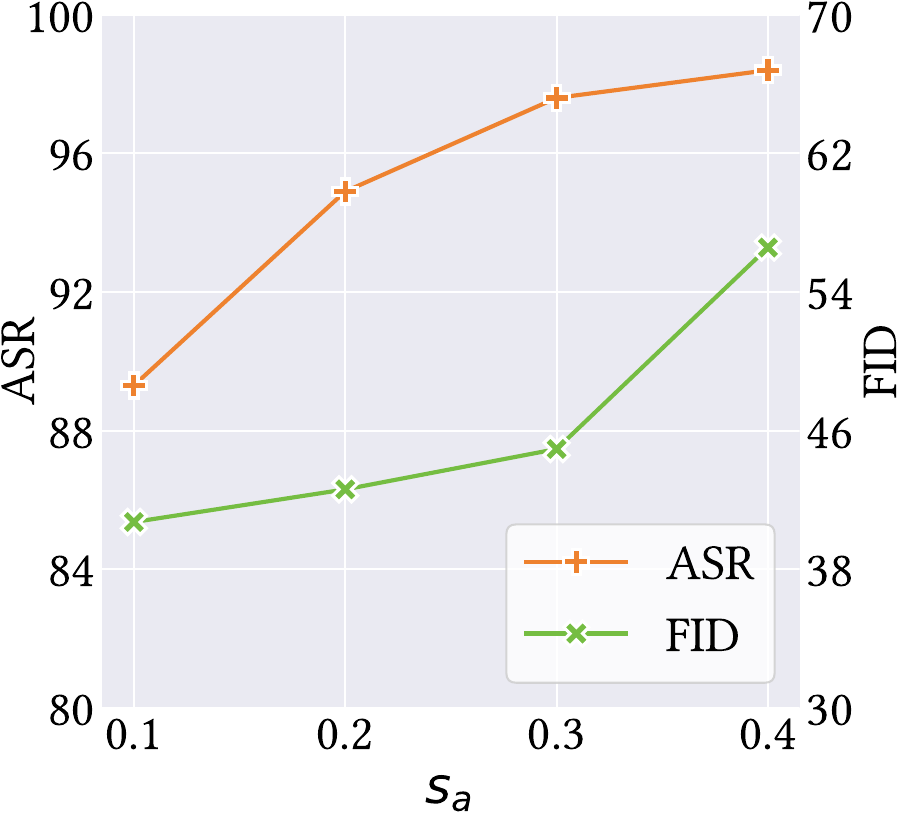}
	\includegraphics[width=0.247\linewidth]{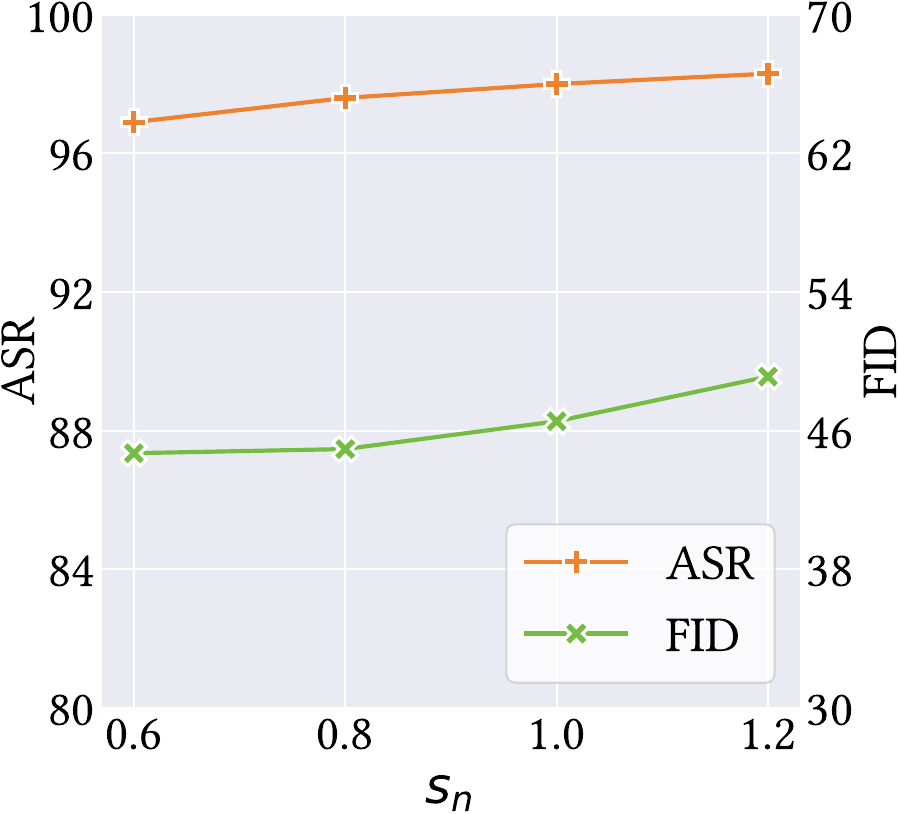}
	\caption{Ablation analysis of ScoreAdv parameters. We selected ResNet-50 as the target model and employed ASR and FID as the evaluation metrics to assess the attack effectiveness and generated image quality, respectively.}
	\label{parameters}
\end{figure*}

\begin{table}[h]
\caption{Image quality comparison of various methods on ImageNet dataset. The best result is \textbf{bolded} and the second best result is \underline{underlined}.}
\label{quality}
\begin{tabularx}{\linewidth}{lXXXX}\midrule
Attacks     & FID$\downarrow$ & LPIPS$\downarrow$ & PSNR$\uparrow$ & SSIM$\uparrow$ \\ \midrule
PGD         & 72.740 & 0.276 & 23.867 & 0.6541     \\
DI$^2$-FGSM & 71.650 & 0.198 & 26.388 & 0.7599     \\
S$^2$I-FGSM & 51.022 & 0.158 & 29.808 & 0.8296     \\
cAdv        & 69.593 & 0.246 & 24.113 & 0.7059     \\
NCF         & 70.857 & 0.355 & 14.994 & 0.6087     \\
DiffAttack  & 48.639 & 0.169 & 27.450 & 0.6546     \\
AdvDiffuser & \underline{45.399} & \underline{0.137} & \underline{30.146} & \underline{0.8302}     \\ \midrule
ScoreAdv    & \textbf{44.932} & \textbf{0.124} & \textbf{30.817} & \textbf{0.8319}     \\ \midrule
\end{tabularx}
\end{table}

ScoreAdv consistently produces the most natural and coherent adversarial images. In Figure \ref{visual}, it best preserves perceptual fidelity and captures high-frequency details, enhancing its realism. Table \ref{quality} presents the quantitative comparison of image quality. ScoreAdv achieves the lowest FID, indicating better preservation of the reference image distribution. Higher PSNR and SSIM reflect minimal, well-placed perturbations, while lower LPIPS confirms better alignment with human perception.

\paragraph{Robustness Against Defense Methods.}
To evaluate the robustness of ScoreAdv, we conducted a series of adversarial defense experiments. Specifically, AdvProp \cite{xie2020-adversarial}, R\&P \cite{xie2017-mitigating}, RandS \cite{cohen2019-certified}, Bit-Red \cite{xu2018-feature}, Adv-Inc-v3 \cite{alexey2018-adversarial}, and IncRes-v2$_{ens}$ \cite{alex2018-ensemble} apply defense mechanisms or train robust models before attack. SR \cite{mustafa2020-image}, NRP \cite{naseer2020-a}, and DiffPure \cite{nie2022-diffusion} operate by purifying images post-attack to mitigate adversarial effects.

As shown in Table \ref{robustness}, ScoreAdv demonstrates superior robustness compared to previous methods and exhibits slightly better robustness than AdvDiff. This enhanced performance may be attributed to both methods that utilize the inherent distribution-shifting property of diffusion models. Moreover, the dynamic interplay of iterative denoising and adversarial perturbation injection inadvertently contributes to improving the robustness of the generated images.

\paragraph{Computational Analysis}

ScoreAdv is designed to be computationally practical, it internalizes adversarial guidance into the diffusion sampling loop and performs all saliency and gradient work on a surrogate model, avoiding costly queries or retraining of the target. Importantly, ScoreAdv's end-to-end runtime is comparable to the original diffusion model's inference speed, the additional overhead introduced by our guidance and saliency steps is effectively negligible since most cases use $N=1$. In our setup, the end-to-end average inference time is under 10s per image, roughly 3$\times$ faster than comparable diffusion-based attack baselines like DiffAttack \cite{chen2025-diffusion} and AdvDiffuser \cite{chen2023-advdiffuser}, which require over 30s per image. The dominant cost is the diffusion sampler which scales with $T$; ScoreCAM calls are amortized, so they add only a small overhead.

\paragraph{Additional Results.}
Please see Supplementary Material \ref{section additional} for additional experiments on more SOTA methods and ensemble attacks and ablation studies on saliency maps.

\subsection{Parameter Analysis}

In Figure \ref{parameters}, we conduct a comprehensive parameter analysis on diffusion steps $T^*$, noise optimization times $N$, adversarial guidance scale $s_a$, and noise optimization scale $s_n$.

\paragraph{Noise Addition Steps $T^*$.} $T^*$ denotes the number of diffusion steps where adversarial perturbations are applied, specifically during the final $T^*$ steps. As shown, increasing $T^*$ improves ASR but may degrade image quality due to accumulated noise. Our $T^*$ balances ASR and visual fidelity.

\paragraph{Noise Optimization Times $N$.} A larger $N$ enables further optimization of the initial noise, improving adversarial effectiveness and ASR, similar to increasing iterations in GANs. However, excessive $N$ may distort the noise too much, degrading image quality. As shown in the figure, a small $N$ already yields high ASR, so we choose a moderate value to balance performance and visual fidelity.

\paragraph{Adversarial Guidance Scale $s_a$ And Noise Optimization Scale $s_n$.} 
In the figure, increasing $s_a$ improves attack performance by amplifying perturbation strength and enhancing guidance toward the target class. However, overly large $s_a$ degrades image quality. The effect of $s_n$ follows a similar trend but has less impact, as it only applies when $N > 1$.

\section{Conclusion}

In this paper, we propose ScoreAdv, an interpretable method for generating natural UAEs by leveraging the distribution shift in diffusion models. It uses a pretrained diffusion model with adversarial guidance injected during denoising, and integrates ScoreCAM to incorporate visual cues from a reference image. Experiments show that ScoreAdv achieves SOTA performance in both black-box and white-box attacks on classification and recognition tasks. Crucially, it manages to maintain exceptionally high image quality and exhibits strong robustness against defensive mechanisms, all while remaining computationally efficient. Moreover, since ScoreAdv only relies on the gradient of a generic scoring function, its formulation is naturally extensible to vision-language models by substituting $f$ with multimodal encoders. This flexibility further highlights its generality beyond conventional image classifiers.

{
    \small
    \bibliographystyle{ieeenat_fullname}
    \bibliography{main}
}


\clearpage

\setcounter{page}{1}
\maketitlesupplementary

\section{Theoretical Derivation}\label{Theoretical Derivation}

\begin{table*}[h!]
\caption{Attack performance comparison of various methods on image recognition models. We report ASR (\%) and transferability. S. represents the surrogate model, while T. represents the target model. In cases where the surrogate model and the target model are identical, the scenario is classified as a white-box attack, with the background highlighted in gray for clarity. The best result is \textbf{bolded} and the second best result is \underline{underlined}.}
\label{recognition performance}
\begin{tabularx}{\textwidth}{XXXXXXXX}
\midrule
\multirow{2}{*}{\diagbox[innerwidth=4em, height=2.3em]{S.}{T.}}       & \multirow{2}{*}{Attacks} & \multicolumn{3}{c}{LFW}      & \multicolumn{3}{c}{CebebA-HQ} \\ \cmidrule(r){3-5} \cmidrule(r){6-8}
                            &                          & IR152 & FaceNet & MobileFace & IR152  & FaceNet & MobileFace \\ \midrule
\multirow{8}{*}{IR152}      & DI                       & \cellcolor{lightgray!60}\underline{95.4}  & 21.7    & 14.4       & \cellcolor{lightgray!60}87.9   & \underline{27.6}    & 27.0       \\
                            & VMI                      & \cellcolor{lightgray!60}92.7  & 23.9    & 18.7       & \cellcolor{lightgray!60}92.2   & 26.4    & 27.7       \\
                            & SSA                      & \cellcolor{lightgray!60}78.8  & 21.9    & \underline{24.0}       & \cellcolor{lightgray!60}83.8   & 23.3    & \underline{30.7}       \\
                            & DFANet                   & \cellcolor{lightgray!60}\textbf{98.9}  & 15.5    & 12.5       & \cellcolor{lightgray!60}\textbf{98.9}   & 21.5    & 22.4       \\
                            & SIA                      & \cellcolor{lightgray!60}81.7  & \underline{26.3}    & 19.6       & \cellcolor{lightgray!60}78.4   & 26.9    & 30.4       \\
                            & BSR                      & \cellcolor{lightgray!60}52.4  & 14.7    & 7.3        & \cellcolor{lightgray!60}48.5   & 17.3    & 15.0       \\
                            & BPFA                     & \cellcolor{lightgray!60}92.6  & 12.9    & 9.2        & \cellcolor{lightgray!60}90.4   & 15.8    & 16.4       \\ \cmidrule{2-8}  
                            & ScoreAdv                 & \cellcolor{lightgray!60}\textbf{98.9}      & \textbf{83.9}        & \textbf{97.5}           & \underline{95.0}   & \cellcolor{lightgray!60}\textbf{55.0}    & \textbf{84.0}       \\ \midrule
\multirow{8}{*}{FaceNet}    & DI                       & 18.6  & \cellcolor{lightgray!60}\textbf{99.8}    & 18.5       & 22.8   & \cellcolor{lightgray!60}99.4    & 22.9       \\
                            & VMI                      & 24.4  & \cellcolor{lightgray!60}\textbf{99.8}    & 20.7       & 26.4   & \cellcolor{lightgray!60}99.3    & 25.7       \\
                            & SSA                      & 21.6  & \cellcolor{lightgray!60}97.5    & \underline{30.8}       & 23.5   & \cellcolor{lightgray!60}96.9    & \underline{36.1}       \\
                            & DFANet                   & 12.1  & \cellcolor{lightgray!60}\textbf{99.8}    & 11.7       & 15.2   & \cellcolor{lightgray!60}99.1    & 19.7       \\
                            & SIA                      & \underline{29.1}  & \cellcolor{lightgray!60}\underline{99.5}    & 26.2       & 29.3   & \cellcolor{lightgray!60}\underline{99.4}    & 35.7       \\
                            & BSR                      & 28.6  & \cellcolor{lightgray!60}98.6    & 25.9       & \underline{27.8}   & \cellcolor{lightgray!60}98.8    & 34.0       \\
                            & BPFA                     & 17.3  & \cellcolor{lightgray!60}98.6    & 14.7       & 20.0   & \cellcolor{lightgray!60}99.0    & 21.7       \\ \cmidrule{2-8}  
                            & ScoreAdv                 & \textbf{79.7}      & \cellcolor{lightgray!60} 98.9       & \textbf{89.4}           & \textbf{35.8}       & \cellcolor{lightgray!60}\textbf{99.8}        & \textbf{61.9}           \\ \midrule
\multirow{8}{*}{MobileFace} & DI                       & \underline{18.4}  & \underline{32.9}    & \cellcolor{lightgray!60}99.2       & 25.0   & \underline{32.0}    & \cellcolor{lightgray!60}95.2       \\
                            & VMI                      & 13.6  & 20.2    & \cellcolor{lightgray!60}\underline{99.7}       & 20.2   & 19.6    & \cellcolor{lightgray!60}98.2       \\
                            & SSA                      & 13.8  & 19.6    & \cellcolor{lightgray!60}98.3       & 22.0   & 21.0    & \cellcolor{lightgray!60}93.3       \\
                            & DFANet                   & 7.0   & 11.9    & \cellcolor{lightgray!60}99.6       & 10.7   & 12.6    & \cellcolor{lightgray!60}\underline{99.1}       \\
                            & SIA                      & 15.7  & 26.7    & \cellcolor{lightgray!60}98.4       & \underline{25.4}   & 25.9    & \cellcolor{lightgray!60}96.3       \\
                            & BSR                      & 5.4   & 9.5     & \cellcolor{lightgray!60}84.9       & 10.8   & 11.9    & \cellcolor{lightgray!60}77.6       \\
                            & BPFA                     & 15.8  & 17.8    & \cellcolor{lightgray!60}97.7       & 20.7   & 17.5    & \cellcolor{lightgray!60}96.2       \\ \cmidrule{2-8}  
                            & ScoreAdv                 & \textbf{84.9}      & \textbf{81.5}        & \cellcolor{lightgray!60} \textbf{100.0}          & \textbf{48.1}   & \textbf{40.0}    & \cellcolor{lightgray!60}\textbf{100.0}      \\ \midrule
\end{tabularx}
\end{table*}

\subsection{Derivation of equation \eqref{adversarial guidance}}\label{eq8derivation}

According to the Bayes's theorem $P(A B)=P(A) P(B | A)$ and $P(A B C)=P(A) P(B | A) P(C | A B)$, we want to sample $\tilde{\boldsymbol{x}}_{t-1}$ based on $\bar{\boldsymbol{x}}_{t-1}$ following the equation.
\begin{equation}\label{eq20}
\begin{aligned}
& \quad p_{\theta}\left(\tilde{\boldsymbol{x}}_{t-1} | \bar{\boldsymbol{x}}_{t-1}, y_{tar}\right) \\ & =\frac{p_{\theta}\left(\tilde{\boldsymbol{x}}_{t-1}, \bar{\boldsymbol{x}}_{t-1}, y_{tar}\right)}{p_{\theta}\left(\bar{\boldsymbol{x}}_{t-1}, y_{tar}\right)} \\
& =\frac{p_{\theta}\left(y_{tar} | \tilde{\boldsymbol{x}}_{t-1}, \bar{\boldsymbol{x}}_{t-1}\right) \cdot p_{\theta}\left(\tilde{\boldsymbol{x}}_{t-1} | \bar{\boldsymbol{x}}_{t-1}\right) \cdot p_{\theta}\left(\bar{\boldsymbol{x}}_{t-1}\right)}{p_{\theta}\left(y_{tar} | \bar{\boldsymbol{x}}_{t-1}\right) \cdot p_{\theta}\left(\bar{\boldsymbol{x}}_{t-1}\right)} \\
& =\frac{p_{\theta}\left(y_{tar} | \tilde{\boldsymbol{x}}_{t-1}, \bar{\boldsymbol{x}}_{t-1}\right) \cdot p_{\theta}\left(\tilde{\boldsymbol{x}}_{t-1} | \bar{\boldsymbol{x}}_{t-1}\right)}{p_{\theta}\left(y_{tar} | \bar{\boldsymbol{x}}_{t-1}\right)} \\
& =\frac{p_{\theta}\left(y_{tar}, \tilde{\boldsymbol{x}}_{t-1}, \bar{\boldsymbol{x}}_{t-1}\right)}{p_{\theta}\left(\tilde{\boldsymbol{x}}_{t-1}, \bar{\boldsymbol{x}}_{t-1}\right)} \cdot \frac{p_{\theta}\left(\tilde{\boldsymbol{x}}_{t-1} | \bar{\boldsymbol{x}}_{t-1}\right)}{p_{\theta}\left(y_{tar} | \bar{\boldsymbol{x}}_{t-1}\right)} \\
& =\frac{p_{\theta}\left(\bar{\boldsymbol{x}}_{t-1} | \tilde{\boldsymbol{x}}_{t-1}, y_{tar}\right) \cdot p_{\theta}\left(y_{tar} | \tilde{\boldsymbol{x}}_{t-1}\right)}{p_{\theta}\left(\bar{\boldsymbol{x}}_{t-1} | \tilde{\boldsymbol{x}}_{t-1}\right)} \cdot \frac{p_{\theta}\left(\tilde{\boldsymbol{x}}_{t-1} | \bar{\boldsymbol{x}}_{t-1}\right)}{p_{\theta}\left(y_{tar} | \bar{\boldsymbol{x}}_{t-1}\right)}
\end{aligned}
\end{equation}

Owing to the Markov property of the forward process, we have $p_{\theta}\left(\bar{\boldsymbol{x}}_{t-1} | \tilde{\boldsymbol{x}}_{t-1}, y_{tar}\right)=p_{\theta}\left(\bar{\boldsymbol{x}}_{t-1} | \tilde{\boldsymbol{x}}_{t-1}\right)$. Thus equation \eqref{eq20} can be simplified to:
\begin{equation}
p_{\theta}\left(\tilde{\boldsymbol{x}}_{t-1} | \bar{\boldsymbol{x}}_{t-1}, y_{tar}\right)=\frac{p_{\theta}\left(\tilde{\boldsymbol{x}}_{t-1} | \bar{\boldsymbol{x}}_{t-1}\right) \cdot p_{\theta}\left(y_{tar} | \tilde{\boldsymbol{x}}_{t-1}\right)}{p_{\theta}\left(y_{tar} | \bar{\boldsymbol{x}}_{t-1}\right)}
\end{equation}

We express it in exponential form.
\begin{equation}\label{eq22}
\begin{aligned}
& p_{\theta}\left(\tilde{\boldsymbol{x}}_{t-1} | \bar{\boldsymbol{x}}_{t-1}, y_{tar}\right) \\
&= p_{\theta}\left(\tilde{\boldsymbol{x}}_{t-1} | \bar{\boldsymbol{x}}_{t-1}\right) \cdot e^{\log p_{\theta}\left(y_{tar} | \tilde{\boldsymbol{x}}_{t-1}\right)-\log p_{\theta}\left(y_{tar} | \bar{\boldsymbol{x}}_{t-1}\right)}
\end{aligned}
\end{equation}

We perform a Taylor expansion of $\log p_{\theta}\left(y_{tar} | \tilde{\boldsymbol{x}}_{t-1}\right)$ around the mean of the sampling distribution, denoted as $\tilde{\boldsymbol{x}}_{t-1} = \mu\left(\bar{\boldsymbol{x}}_{t-1}\right)$, where $\mu$ represents the predicted mean of the Gaussian noise during the inference process (denote $\mu\left(\bar{\boldsymbol{x}}_{t-1}\right)$ as $\boldsymbol{\mu}$ for concise).
\begin{equation}\label{eq23}
\begin{aligned}
& \log p_{\theta}\left(y_{tar} | \tilde{\boldsymbol{x}}_{t-1}\right) \\ & \approx \log p_{\theta}\left(y_{tar} | \boldsymbol{\mu}\right)+\left(\tilde{\boldsymbol{x}}_{t-1}-\boldsymbol{\mu}\right)\left[\nabla_{\boldsymbol{\mu}} \log p_{\theta}\left(y_{tar} | \boldsymbol{\mu}\right)\right] \\
& =\left(\tilde{\boldsymbol{x}}_{t-1}-\boldsymbol{\mu}\right) \boldsymbol{g}+C_{1}
\end{aligned}
\end{equation}

\noindent{where $\boldsymbol{g} = \nabla_{\boldsymbol{\mu}} \log p_{\theta}\left(y_{tar} | \boldsymbol{\mu}\right)$, $C_{1}$ is a constant.}

Assume logarithmic probability density function of $p_{\theta}\left(\tilde{\boldsymbol{x}}_{t-1} | \bar{\boldsymbol{x}}_{t-1}\right)$ is given by:
\begin{equation}
\log p_{\theta}\left(\tilde{\boldsymbol{x}}_{t-1} | \bar{\boldsymbol{x}}_{t-1}\right)=-\frac{1}{2}\left(\tilde{\boldsymbol{x}}_{t-1}-\boldsymbol{\mu}\right)^{T} \boldsymbol{\sigma}^{-1}_t\left(\tilde{\boldsymbol{x}}_{t-1}-\boldsymbol{\mu}\right)
\end{equation}

Substituting it into the equation \eqref{eq22} and \eqref{eq23} yields:
\begin{equation}\label{eq25}
\begin{aligned}
& \quad \log p_{\theta}\left(\tilde{\boldsymbol{x}}_{t-1} | \bar{\boldsymbol{x}}_{t-1}, y_{tar}\right) \\ & \approx-\frac{1}{2}\left(\tilde{\boldsymbol{x}}_{t-1}-\boldsymbol{\mu}\right)^{T} \boldsymbol{\sigma}^{-1}_t\left(\tilde{\boldsymbol{x}}_{t-1}-\boldsymbol{\mu}\right)+\left(\tilde{\boldsymbol{x}}_{t-1}-\boldsymbol{\mu}\right) \boldsymbol{g} \\
& =-\frac{1}{2}\left[\left(\tilde{\boldsymbol{x}}_{t-1}-\boldsymbol{\mu}\right)^{T} \boldsymbol{\sigma}^{-1}_t\left(\tilde{\boldsymbol{x}}_{t-1}-\boldsymbol{\mu}\right)-2\left(\tilde{\boldsymbol{x}}_{t-1}-\boldsymbol{\mu}\right) \boldsymbol{g}\right] \\
& =-\frac{1}{2}\left[\left(\tilde{\boldsymbol{x}}_{t-1}-(\boldsymbol{\mu}+\boldsymbol{\sigma}_t \boldsymbol{g})\right)^{T} \boldsymbol{\sigma}^{-1}_t\left(\tilde{\boldsymbol{x}}_{t-1}-(\boldsymbol{\mu}+\boldsymbol{\sigma}_t \boldsymbol{g})\right)\right]+C \\
& =\log p(\boldsymbol{z}), \boldsymbol{z} \sim \mathcal{N}(\boldsymbol{\mu}+\boldsymbol{\sigma}_t \boldsymbol{g}, \boldsymbol{\sigma}_t)
\end{aligned}
\end{equation}

Thus, sampling with equation \eqref{eq25} should be:
\begin{equation}
\tilde{\boldsymbol{x}}_{t-1}=\bar{\boldsymbol{x}}_{t-1} + s_a \boldsymbol{\sigma}_t^2 \boldsymbol{g}
\end{equation}

\subsection{Derivation of equation \eqref{recognition}}\label{eq9derivation}

\begin{figure*}[!h]
  \centering
  \includegraphics[width=1.0\linewidth]{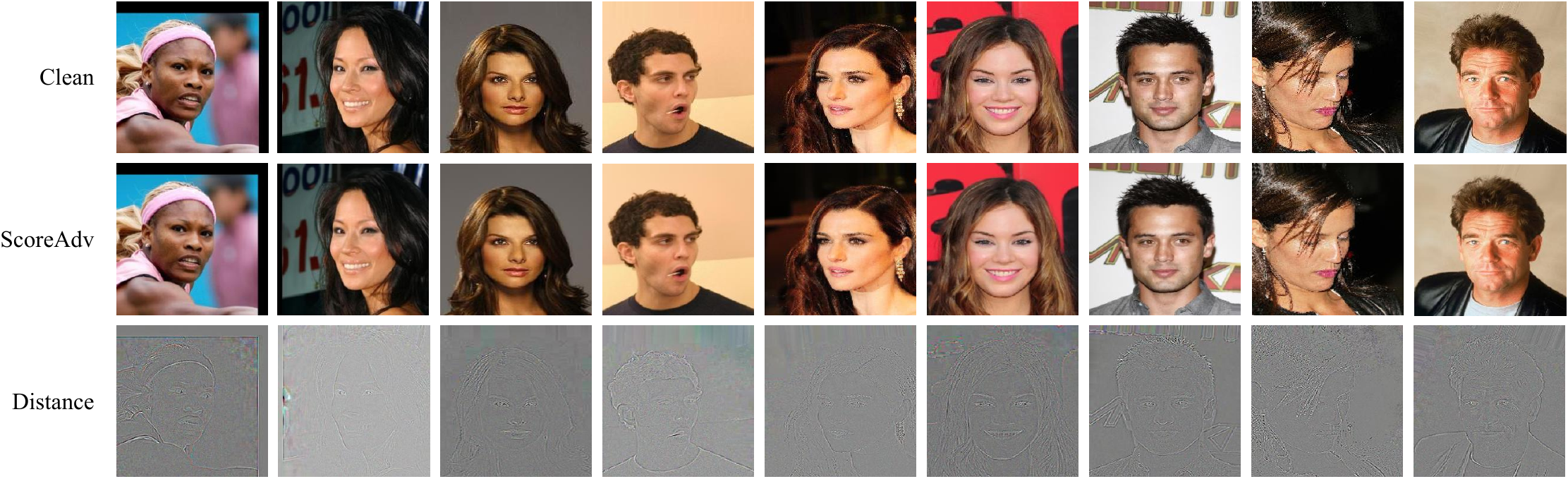}
  \caption{Visualization of adversarial images and perturbations generated by different attack methods. The upper row displays the adversarial images, while the lower row illustrates the corresponding perturbations.}
  \label{visual}
\end{figure*}

The sampling formula for image recognition tasks is similar to that used in image classification tasks, with the primary difference being the replacement of the classification model's gradient, $\nabla_{\bar{\boldsymbol{x}}_{t-1}} \log p_f\left(y_{tar}|\bar{\boldsymbol{x}}_{t-1}\right)$, with the recognition model's gradient, $\nabla_{\hat{\boldsymbol{x}}_0} \log p_f\left(\boldsymbol{x}_{tar}|\hat{\boldsymbol{x}}_0\right)$. All other theoretical derivations remain unchanged.

\subsection{Derivation of equation \eqref{noise optimization}}\label{eq15derivation}

The derivation of equation \eqref{noise optimization} is similar to that of equation \eqref{adversarial guidance}. We perform Taylor expansion around $\boldsymbol{x}_0 = \boldsymbol{x}_T$.
\begin{equation}
\begin{aligned}
& \log p_{\theta}\left(y_{tar} | \boldsymbol{x}_T\right) \\
&\approx \log p_{\theta}\left(y_{tar} | \boldsymbol{x}_0\right)+\left(\boldsymbol{x}_T-\boldsymbol{x}_0\right)\left[\nabla_{\boldsymbol{x}_0} \log p_{\theta}\left(y_{tar} | \boldsymbol{x}_0\right)\right]
\end{aligned}
\end{equation}

From $\boldsymbol{x}_0$ to $\boldsymbol{x}_T$, we employ the forward process of the diffusion model for sampling.
\begin{equation}
p(\boldsymbol{x}_T | \boldsymbol{x}_0)=\mathcal{N}(\boldsymbol{x}_T;\sqrt{\bar{\alpha}_T}\boldsymbol{x}_0,(1-\bar{\alpha}_T)\mathbf{I})
\end{equation}

Thus, the noise optimization process is as follows:
\begin{equation}
\boldsymbol{x}_T= \left(\sqrt{\bar{\alpha}_t} \boldsymbol{x}_0 + \sqrt{1-\bar{\alpha}_t} \boldsymbol{\epsilon}\right) + \bar{\sigma}_T^2 s_a \nabla_{\boldsymbol{x}_0} \log p_f(y_{tar}|\boldsymbol{x}_0)
\end{equation}

\section{Attack Performance on Recognition Systems}\label{face performance}

Table \ref{recognition performance} presents the attack performance of various methods on the recognition model. We select IR152 \cite{he2016-deep}, FaceNet \cite{schroff2015-facenet}, and MobileFace \cite{chen2018-mobilefacenets} as the target recognition model. Our proposed method, ScoreAdv, achieves SOTA performance, surpassing previous approaches such as DI \cite{xie2019-improving}, VMI \cite{wang2021-enhancing}, SSA \cite{long2022-frequency}, DFANet \cite{zhong2021-towards}, SIA \cite{wang2023-structure}, BSR \cite{wang2024-boosting}, and BPFA \cite{zhou2024-improving}. As shown in the table, ScoreAdv significantly outperforms other methods in black-box scenarios and achieves competitive or superior results in most white-box settings. This demonstrates that our method is not only effective in attacking image classification models but also exhibits strong performance when targeting recognition models.

\section{Additional Results}\label{section additional}

\subsection{Comparison With More SOTA Methods}

While we have already compared our method with the most advanced diffusion-based UAE approaches, we further extend our evaluation to include additional SOTA methods, such as TI-FGSM \cite{dong2019-evading} VENOM \cite{zhang2025-venom}, AdvAD \cite{li2024-advad}, and NatADiff \cite{collins2025-natadiff}. As shown in Table \ref{additional}, ScoreAdv consistently achieves superior performance in both attack success rate and image quality. Moreover, although comparing ScoreAdv with ensemble-based attacks is inherently disadvantageous to us, we still include such comparisons. The results demonstrate that our method remains highly competitive against a wide range of baselines.

\begin{table}[h]
\caption{Comparison with additional SOTA methods and ensemble attacks. The evaluation metrics remain consistent with the main paper. STDMI-F refers to S$^2$I-TI-DI-MI-FGSM. The best result is \textbf{bolded} and the second best result is \underline{underlined}.}
\label{additional}
\begin{tabularx}{\linewidth}{llllll}
\toprule
Attacks           & Res50 & Inc-v3 & ViT-B & Mix-B & SSIM  \\ \midrule
TI-FGSM           & 94.8  & 32.5   & \underline{25.9}  & \underline{48.8}  & 0.716 \\
NatADiff          & 96.2  & 18.7   & 6.6   & $-$     & $-$     \\
AdvAD             & \textbf{98.6}  & 37.6   & 14.1  & 31.7  & \textbf{0.887} \\
VENOM             & 97.5  & \underline{45.8}   & 17.2  & 36.0  & 0.715 \\ \midrule
ScoreAdv          & \underline{97.6}  & \textbf{46.7}   & \textbf{40.1}  & \textbf{65.3}  &\underline{0.832} \\ \midrule
STDMI-F           & 99.7  & 51.3   & 44.8  & 64.3  & 0.682 \\ \bottomrule
\end{tabularx}
\end{table}

\subsection{Ablation Study on Saliency Map}

As shown in Table \ref{scorecam}, we compare the proposed ScoreCAM with other interpretability-based saliency methods, including GradCAM \cite{selvaraju2017-gradcam}, and GradCAM++ \cite{chattopadhyay2018-gradcam++}. Experimental results demonstrate that our approach consistently outperforms the others in both attack effectiveness and image quality.

\begin{table}[h]
\caption{Ablation study on different saliency maps. The best result is \textbf{bolded} and the second best result is \underline{underlined}.}
\label{scorecam}
\begin{tabularx}{\linewidth}{lXXXX}
\toprule
Saliency maps & Res50 & Inc-v3 & Mix-B & FID    \\ \midrule
GradCAM       & 92.6  & 43.1   & 63.0  & 56.758 \\
GradCAM++     & 93.1  & 43.5   & 62.8  & 54.629 \\ \midrule
ScoreAdv      & \textbf{97.6}  & \textbf{46.7}   & \textbf{65.3}  & \textbf{44.932} \\ \bottomrule
\end{tabularx}
\end{table}

\end{document}